# An End-to-End Decision-Aware Multi-Scale Attention-Based Model for Explainable Autonomous Driving

Maryam Sadat Hosseini Azad, Shahriar Baradaran Shokouhi, Amir Abbas Hamidi Imani, Shahin Atakishiyev, *Member, IEEE,* and Randy Goebel

***Abstract*—The application of computer vision is gradually increasing across various domains. They employ deep learning models with a black-box nature. Without the ability to explain the behavior of neural networks, especially their decision-making processes, it is not possible to recognize their efficiency, predict system failures, or effectively implement them in real-world applications. Due to the inevitable use of deep learning in fully automated driving systems, many methods have been proposed to explain their behavior; however, they suffer from flawed reasoning and unreliable metrics, which have prevented a comprehensive understanding of complex models in autonomous vehicles and hindered the development of truly reliable systems. In this study, we propose a multi-scale attention-based model in which driving decisions are fed into the reasoning component to provide case-specific explanations for each decision simultaneously. For quantitative evaluation of our model's performance, we employ the F1-score metric, and also proposed a new metric called the Joint F1 score to demonstrate the accurate and reliable performance of the model in terms of Explainable Artificial Intelligence (XAI). In addition to the BDD-OIA dataset, the nu-AR dataset is utilized to further validate the generalization capability and robustness of the proposed network. The results demonstrate the superiority of our reasoning network over the classic and state-of-the-art models.**

## I. Introduction

DEEP Neural Networks (DNNs) have become essential components in computer vision systems and safety-critical applications. However, the high complexity of these networks makes it challenging to understand their internal workings and decision-making processes. In other words, they have an inherently non-transparent and opaque nature, which prevents them from being explainable. The black-box nature of these networks makes it difficult to trust their outputs. This problem is particularly pronounced under real-world conditions, which are substantially more complex than laboratory settings.

Maryam Sadat Hosseini Azad, Shahriar Baradaran Shokouhi, and Amir Abbas Hamidi Imani are with the Department of Electrical Engineering, Iran University of Science and Technology, Tehran, Iran, (e-mail: hosseini_maryam@elec.iust.ac.ir; bshokouhi@iust.ac.ir; Hamidi_a@elec.iust.ac.ir). *(Corresponding author: Maryam Sadat Hosseini Azad).*

Shahin Atakishiyev and Randy Goebel are with the Department of Computing Science, University of Alberta, Edmonton, AB T6G, Canada (e-mail: Shahin.atakishiyev@ualberta.ca; rgoebel@ualberta.ca)

Autonomous driving, as a prominent application, has many complexities that necessitate the use of DNNs. In this regard, if the network is interpretable, system troubleshooting becomes possible at a much higher speed. Consequently, the development speed of self-driving cars would drastically increase. Public trust is another critical challenge in this domain, and the significant impact of explainability on user acceptance has been thoroughly investigated [1, 2]. Moreover, after understanding the internal workings of self-driving decision systems, policymakers can more easily and reliably establish traffic laws [3]. Explainable artificial intelligence (XAI) methods can be presented in three main forms [4]:

- Visual explanations, which include diagrams, graphs, and heatmaps (e.g., SHAP [5])
- Textual justifications that express explanations through natural language [6] or formal representations [7]
- Auditory explanations [8]

Considering that each of these presentation methods is individually incomplete, we employed both visual and textual methods to present our network explanation. While textual explanations using natural language are more understandable to users and contribute to broader acceptance of autonomous driving technology, formal representations are incomprehensible to the general public. Moreover, visual explanations not only provide meaningful insights into model behavior but also help identify weak points and potential areas for improvement. It should be noted that auditory explanations are limited by their sequential nature and the speed at which information can be conveyed.

On the other hand, there are two general approaches in autonomous driving systems: end-to-end and modular pipeline [9]. Autonomous driving requires the cooperation of key sub-modules—including perception [10], localization [11], system management [12], planning [13], and control [14]—to make correct driving decisions [15]. However, providing explanations for individual modules is insufficient to ensure transparency of the overall system [16]. Therefore, in this study, we employed an end-to-end model to provide a comprehensive explanation.

In recent years, with remarkable progress in image understanding and deep learning, researchers have proposed various methods to address the transparency limitations of neural networks. To the best of our knowledge, there are two fundamental points that have not yet been addressed in prior research:
First, in the area of XAI research focused on understanding complex model decisions, which is precisely the challenge addressed in this paper, to explain a particular decision, the explainer needs to first identify what decision the network produced [17-19]. In previous works, the reasoning head essentially provides transparency for a network that is unaware of its own decision [20, 21]. Second, a high accuracy alone is insufficient for trusting a model; the reasons behind the predictions must also be logical [22]. Even highly accurate neural networks may not follow the correct logic for their decisions [23]. Models with high accuracy but incorrect reasoning are compared with models with lower accuracy but more logical reasoning, and users can utilize the explanations to choose models that perform better in the real world, even if their apparent accuracy is lower [24].

In this study, we present a novel model that effectively captures both important fine-grained local details and broader contextual information essential for driving behavior explanation. Besides, it employs a sound reasoning mechanism for generating explanations, where the action outputs are fed into the reasoning module to understand the decision and then generate reasoning for it. Furthermore, by simultaneously learning the driving decisions and rationale behind them, the model leverages the hierarchical relationships between the action and reason modules. As illustrated in Fig. 1, the proposed model generates driving decisions along with both textual and visual explanations to explain the internal workings of the model. Moreover, the model is evaluated using a newly proposed XAI metric designed to evaluate the true performance of the model. In summary, the main contributions of this research are as follows:

- We propose a new explainable decision-making network for autonomous driving: An end-to-end multi-task attention-based deep learning model with Multi-scale Atrous Spatial Pyramid Pooling (MASPP).
- A novel approach proposed in this study involves feeding the output of the driving decisions into the reasoning module, where the reasoning component generates explanations while being aware of the decision made.
- Introducing a new, accurate, and reliable Joint F1 score to evaluate the realistic performance of the model in terms of XAI.
- The superiority of our model has been demonstrated using two standard datasets. The model achieves much higher explanation accuracy, outperforming other classic and state-of-the-art models. Additionally, the effects of different parameters on the model were thoroughly investigated.

## II. Related Works

As image understanding and deep learning networks become more complex, the importance of XAI has increased, particularly in critical applications such as self-driving cars, and many efforts have been made to solve this challenge [25, 26]. To enhance the transparency of end-to-end autonomous driving, the HIIL model was proposed [27], which improves the interpretability of the system by combining an interpretable Bird's Eye View (BEV) mask and steering angle. Another approach is the Simplified Objectification Branches (SOB) structure [28]. This method improves the interpretability by focusing on two key properties - the objectification degree and the simplification degree. The researchers also introduced an interpretable reinforcement learning method that employs a sequential latent environmental model trained simultaneously with the reinforcement learning process [29]. Another set of studies has attempted to provide natural language explanations for the vehicle [30, 31], while some other lines of research has focused on visual explanations [32, 33].

We propose an end-to-end model that increases the accuracy of explainability, particularly for complex decisions, which enables more precise mapping between decisions and their underlying reasons compared to previous approaches. Our goal is to provide more comprehensive and accurate explainability for users, regulators, and engineers.

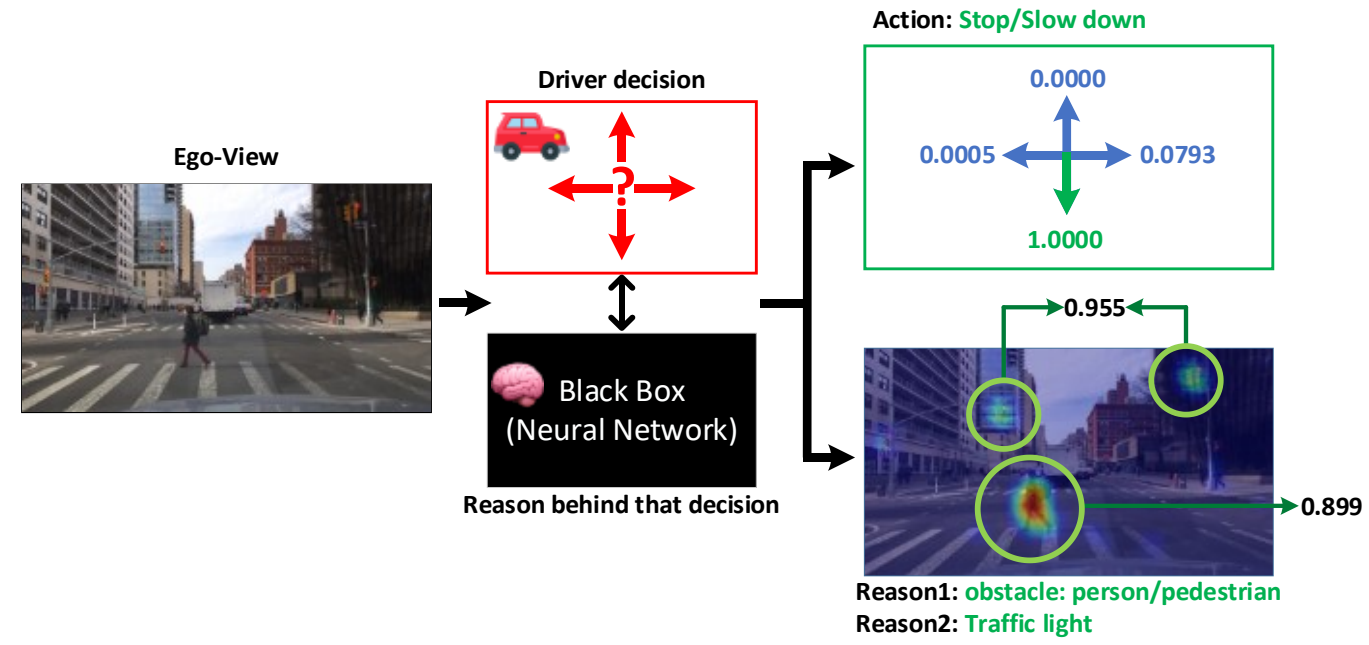

**Fig. 1.** Overview of the problem and its solution investigated in this study. The problem is to understand the driving decision of the ego-vehicle from a dashboard camera image and explain why that decision was made by this black-box model. The output includes the predicted next decision of the vehicle, and textual along with visual reasons behind the prediction. The decision confidence scores are shown as numbers between 0 and 1.

## III. The Proposed Model

### A. Problem Definition

As shown in Fig. 1, the problem is to predict the next action of the self-driving car and provide the reason for that decision from dash-cam images. The mathematical representation can be expressed as:

$$(\text{Action}, \text{Reason}) \in \{0,1\}^{n_a} \times \{0,1\}^{n_r}, \quad (1)$$

This represents the total number of possible action and explanation states, where $n_a$ denotes the number of action categories, and $n_r$ is the number of reason categories. Each driving scene can encompass one or more actions and their associated explanations.

### *B. An End-to-End Attention-Based Deep Learning Model with Multi-Scale Atrous Spatial Pyramid Pooling*

The architecture of our proposed model is shown in Fig. 2. We propose a novel action-explanation network that integrates semantic segmentation with behavioral prediction. According to the architecture, the input to the action-explanation network consists of dashboard camera images, which are fed into ResNet50 [34] along with classification annotations. We also used the pre-trained weights of the DeepLabV3 [35] network, which was trained on the BDD-10K [36] dataset. To address the challenge of multi-scale objects in a driving scene, the MASPP [37] module is utilized to combine multi-scale features from different layers of a deep neural network, incorporating features extracted from layers 2, 3, and 4 of ResNet50 as input. It performs ASPP [38] operations on the fourth layer (highest level features) using atrous rates of 12, 24, and 36. To improve feature extraction at multiple scales, convolution operations with a filter size of 3 are performed using dilation rates of 1 and 3 on the third and second layers, respectively. As a result, the outputs of layers 2, 3, and 4, which may have different dimensions, are adjusted to a uniform size using convolutional layers or pooling methods. The outputs of the three layers are concatenated, and a projection layer is applied to them, which reduces the number of output channels from 2048 to 512 using a 1×1 convolution. This dimensionality reduction improves computational efficiency and decreases memory requirements while maintaining the performance of the model. Consequently, a combinational feature map is produced, which can be expressed as follows:

$$F = f^{1\times1}(F_1;\ F_2;\ \ldots;\ F_7), \tag{2}$$

where $F_i$ represents the output of convolutions with different atrous rates, and $f^{1\times1}$ is the convolution operation with a filter size of 1.

The intermediate feature maps are directed to the DCA-CBAM-ResNet50 attention module, which applies both channel and spatial attention mechanisms to extract important features. This attention module is a hybrid model that combines CBAM [39] and DCANet [40] architectures to enhance attentional features in convolutional neural networks with low computational overhead. Hence, a self-attention mechanism is employed that improves neural network classification results by amplifying relevant features and removing noise, while simultaneously establishing connections between attention blocks. These connections, formulated in Equation 3, transfer attention information from previous blocks to subsequent ones. This approach facilitates the gradual learning of attention mechanisms and prevents large changes in attentional characteristics:

$$Connection\ Function = Y(\alpha G, \beta \tilde{T}), \tag{3}$$

where Y(.) represents the connection function, α and β are trainable parameters, G is the output, and $\tilde{T}$ denotes the attention weights computed from the preceding attention block.

The combined feature map F, generated in the previous step, is sequentially processed through a one-dimensional channel attention module and then a two-dimensional spatial attention module. Attention connections are considered separately for each dimension, which reduces the number of parameters and computational complexity. Additionally, each dimension focuses on its specific features. The fundamental formulas of the attention are as follows. Each attention block receives both the input feature map and the attention information from the previous block, as described in Equation 3:

$$F' = M_c(F, Pre_att) \otimes F, \tag{4}$$

$$F'' = M_s(F', Pre_att) \otimes F', \tag{5}$$

where $F'$ represents the channel attention output and $F''$ denotes the final refined output. $M_c$ and $M_s$ are the channel and spatial attention maps, respectively. $Pre_att$ indicates the attention information of the previous blocks. The symbol ⊗ shows element-wise multiplication.

In channel attention, the output $F$ is first fed into this module. If attention information from the previous layer is available, it is also imported into the module. When the dimensions of $Pre_att$ do not match the input $F$, a 1×1 convolution is performed on it. Both adaptive max-pooling and adaptive average-pooling operations are applied to both inputs. Then, the outputs of the adaptive average-pooling operations are concatenated; similarly, the outputs from the adaptive max-pooling operations are concatenated. After passing through a 1×1 convolution layer, which merges the two incoming channels into one, it enters a shared MLP (Multi-Layer Perceptron) that uses a reduction ratio number to decrease the number of parameters. The shared MLP structure consists of two fully connected layers. First, MLP (C to C/r) reduces the number of features from C to C/r (where r is the reduction ratio). Secondly, another MLP restores the number of features to C. Note that each of the two MLPs has one hidden layer. Then, the outputs are combined through an element-wise summation, which uses a direct connection [40]:

$$Y(\alpha G_i, \beta \tilde{T}_i) = \alpha G_i + \beta \tilde{T}_i, \tag{6}$$

where $i$ denotes the index of a feature.

Finally, the channel feature maps fed into a sigmoid activation function to select the most important channels:

$$M_c(F, Pre_att) = \sigma\Big(MLP\Big(f^{1\times1}\Big(AvgPool(F); AvgPool\big(f^{1\times1}(Pre_att)\big)\Big)\Big) + MLP\Big(f^{1\times1}\Big(MaxPool(F); MaxPool\big(f^{1\times1}(Pre_att)\big)\Big)\Big)\Big), \tag{7}$$

where σ is the sigmoid activation function.

The output is multiplied element-wise by the input F and the result is then passed through the spatial attention module. The module receives $F'$ and previous attention information (if available) as inputs. If required, functional average pooling is performed to match the size of the previous information with $F'$. Then, both adaptive max-pooling and adaptive average-pooling operations are applied separately to each input. The outputs of each input are concatenated along the channel dimension. Finally, a weighted summation with coefficients p1 and p2 is performed according to the weighted connection scheme [40]:

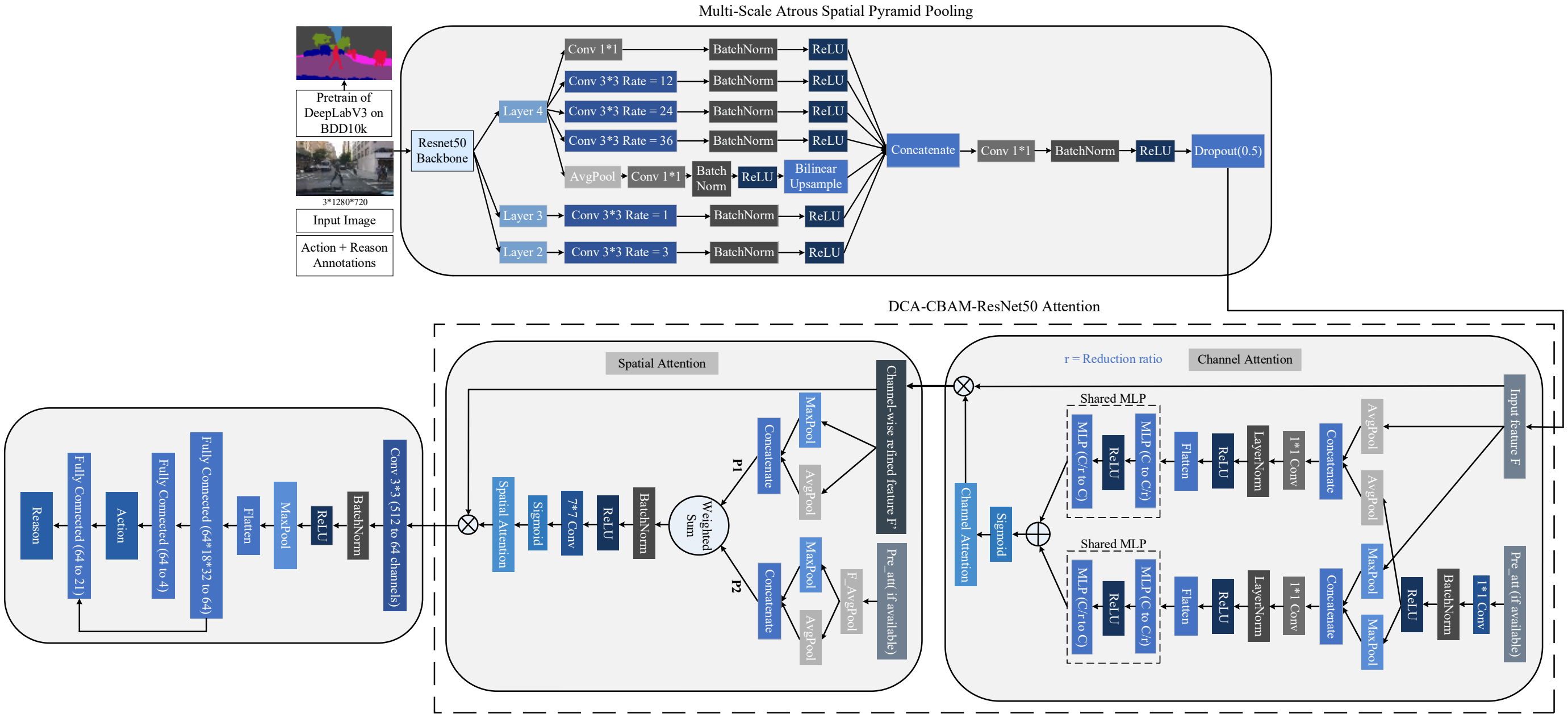

**Fig. 2.** The architecture of our proposed model. The structure consists of three main parts. The first part is a feature extractor using semantic segmentation, ResNet50, and a multi-scale feature extraction module for processing input RGB images. The results are then fed to the attention module, which has both channel and spatial attention with connections between the attention blocks. In the third part, the feature dimensions are reduced and directed to the action and reason heads. Also, the sigmoid probabilities generated for decisions are concatenated with feature dimensions and then fed into a fully connected layer to provide reasons.

$$\Upsilon(\alpha G_i, \beta\tilde{T}_i) = \frac{|\alpha G_i|^2}{\alpha G_i + \beta\tilde{T}_i} + \frac{|\beta\tilde{T}_i|^2}{\alpha G_i + \beta\tilde{T}_i}, \tag{8}$$

A 7×7 convolution is employed because a larger receptive field is needed to achieve higher accuracy. After passing through the sigmoid function, the most important spatial locations are selected:

$$M_s(F, Pre_att) = \sigma\left(f^{7\times 7}\left(P1([AvgPool(F); MaxPool(F)]) + P2\left(\begin{bmatrix} AvgPool(F_AvgPool(Pre_att)) \\ ; MaxPool(F_AvgPool(Pre_att)) \end{bmatrix}\right)\right)\right), \tag{9}$$

The output containing the most important extracted feature maps is multiplied element-wise with the input of $F'$, and the result is fed into the next module. Subsequently, the output sequentially passes through a 3×3 convolution, adaptive max-pooling, as well as fully connected layers until action classification is completed, and the images are mapped to the relevant actions:

$$Action = MLP\left(MLP\left(Maxpool\left(f^{3\times 3}(M_s(F, Pre_att))\right)\right)\right), \tag{10}$$

Then, we feed the action outputs in the sigmoid probability format into the reasoning module, where they are concatenated with the extracted features obtained before the action fully connected layer. Let $X$ be the input space where $X \subseteq \mathbb{R}^{H\times W\times C}$, $Y_a$ be the action output space with 4 dimensions where $Y_a = \{0,1\}^{n_a}$, and $Y_r$ be the reason output space with 21 dimensions where $Y_r = \{0,1\}^{n_r}$. The joint output space for multi-task learning is defined as:

$$Y = Y_a \times Y_r, \tag{11}$$

We define a non-linear function $M$ (model) as $M: X \to Y_a \times Y_r$, which generates both action and reason outputs. The model consists of two components where the reason output depends on the action output:

$$M_a: X \to Y_a\text{: for the action head,}$$
$$M_r: X \times Y_a \to Y_r\text{: for the reasoning head,} \tag{12}$$

The network is defined as follows:

$$M(x) = (M_a(x), M_r(x, M_a(x))), \tag{13}$$

We express a decision function D to convert soft outputs into deterministic decisions:

$$D: Y_a \times Y_r \to \{0,1\}^{k_a} \times \{0,1\}^{k_r},$$
$$\text{where } k_a = 4 \text{ and } k_r = 21, \tag{14}$$

For the action and reasoning outputs:

$$D_a(y_a) = \{i: \sigma(y_{ai}) > \theta\},$$
$$D_r(y_r) = \{i: \sigma(y_{ri}) > \theta\}, \tag{15}$$

where $\sigma$ is the sigmoid function and $\theta$ denotes the decision threshold, which is considered 0.5 in this study.

For each input instance $x \in X$, providing an explanation is only meaningful when the definitive decision of the model is well-defined:

$$\forall x \in X: \exists E(x) \leftrightarrow D(M(x)) \neq \phi, \tag{16}$$

where $E(x)$ is the explanatory function for the sample $x$. The explanation function is defined as:

$$E: X \times (Y_a \times Y_r) \to \mathbb{R}^{n\times m}, \tag{17}$$

where $n \times m$ represents the explanation map dimensions.

The condition for the existence of an explanation is expressed as follows:

$$E(x, (y_a, y_r)) = \{\varphi(x, (y_a, y_r)); \text{ if } D(y_a, y_r) \text{ is defined undefined; otherwise}\}, \tag{18}$$

where $\varphi$ is the descriptive feature extraction function.

Since the networks are identical up to the final layer, the only information that the reason head did not observe from the action pathway was the final layer output, which we subsequently incorporated. Note that we employed concatenation for the implementation. Thus, the reason classification is also completed, and the reasons behind the actions are revealed:

$$Reason = MLP\left(MLP\left(Maxpool\left(f^{3\times3}\left(M_s(F, Pre_att)\right)\right)\right); P_A1; P_A2; P_A3; P_A4\right), \quad (19)$$

where $P_A$ is the sigmoid probabilities of the action outputs.

### C. Joint F1 Score

The critical metric that must be considered is one that evaluates whether correct decisions are made for the right reasons. The more accurate this metric is, the greater trust we can place in the model outputs.

TABLE I
DETAILS OF CORRECT ACTION-CORRECT REASON COMBINATIONS IN THE BDD-OIA DATASET. IT CONSISTS OF THIRTY-THREE DISTINCT PAIRS.

| Action Categories | Reason Categories |
|---|---|
| Move forward | Follow traffic<br>Road is clear<br>Traffic light is green<br>Obstacles on the left lane<br>No lane on the left<br>Solid line on the left<br>Obstacles on the right lane<br>No lane on the right<br>Solid line on the right |
| stop | Obstacle: car<br>Obstacle: person/pedestrian<br>Obstacle: rider<br>Obstacle: others<br>Traffic light<br>Traffic sign<br>Obstacles on the left lane<br>No lane on the left<br>Solid line on the left<br>Obstacles on the right lane<br>No lane on the right<br>Solid line on the right |
| Turn left | Front car turning left<br>On the left-turn lane<br>Traffic light allows<br>Obstacles on the right lane<br>No lane on the right<br>Solid line on the right |
| Turn right | Front car turning right<br>On the right-turn lane<br>Traffic light allows<br>Obstacles on the left lane<br>No lane on the left<br>Solid line on the left |

Accordingly, we introduce a new metric that calculates an F1 score based on whether both the action and reason are correct, which we call the Joint F1 score. There are four action categories and twenty-one reason categories in the BDD-Object-Induced Action (BDD-OIA) [20] dataset. Logically, each action category could be associated with some of the reason categories, but not all of them. It should be noted that when both "turn right" and "turn left" actions are active simultaneously, the reasons related to not being able to turn in both directions are logically implausible. Ultimately, there are thirty-three different cases that can be considered correct action-correct reason combinations as shown in Tab. I. The formula for the introduced metric is shown in Equations 20 and 21. The overall Joint F1 score calculates the average F1 score for each instance, where valid action-reason combinations are considered:

$$F1_{overall}^{joint} = \frac{1}{|N|}\sum_{j=1}^{|N|} F1(\widehat{J_j}, J_j), \quad (20)$$

where N is the total number of samples, $J_j$ represents the valid action-reason combinations for sample j, and $\widehat{J_j}$ represents the predicted action-reason combinations for sample j.

The mean Joint F1 score calculates the average F1 score for each class (each valid combination):

$$F1_{mean}^{joint} = \frac{1}{|n_{valid}|}\sum_{i=1}^{|n_{valid}|} F1_i^{joint}, \quad (21)$$

$n_{valid}$ denotes the total number of valid action-reason combinations.

If a category has exactly zero ground-truth samples in the dataset, F1 = 0 is assigned to that class and included in the macro-averaging calculation. This metric shows the real performance of the black box model and if it achieves high percentages, it is possible to implement the model in real-world scenarios.

At the end, we used Gradient-weighted Class Activation Mapping (Grad-CAM) [41] to generate visual explanations. It is a gradient-based class-discriminative localization technique that generates pixel-level explanations to explain how CNN-based networks make decisions. Grad-CAM gradients are computed as a weighted sum of the predicted action and reason logits, with equal weights of 0.5 for each component. This model is an interpretable method that identifies key regions in an image that significantly contribute to the decision-making process of the model. Note that this method passes the tests proposed in the [42, 43] and provides true interpretability.

## IV. EXPERIMENT AND EVALUATION

### A. Training Details

We used the pre-trained weights of DeepLabV3 [35] on BDD-10K [36] to completely capture and understand the scene at a pixel-wise level. Then, we fine-tuned DeepLabV3 on the BDD-OIA [20] dataset. Because there was not complete overlap between these datasets and BDD-10K, slight adaptations were necessary.

Since actions and reasons are predicted in parallel, our task employs multi-task learning [44], and consequently, its loss function is also a multi-task loss function:

$$L\,(action, reason) = L_{action} + \lambda L_{reason} = \sum_{i=1}^{4} L\left[\widehat{A_i}, A_i\right] + \lambda\sum_{j=1}^{21} L\left[\widehat{R_j}, R_j\right], \quad (22)$$

where $A_i$ and $R_j$ are the human-labeled action and reason of the sample $i$ and $j$, respectively. $\widehat{A_i}$ as well as $\widehat{R_j}$ denote predicted action and reason for sample $i$ and $j$, respectively. $\lambda$ is the relative importance value between action and reasoning.

Given that there are 4 classes for action and 21 categories for reason, action and reason can be described as binary vectors, for example, A = [0, 1, 0, 1]$^T$, which are considered multi-label annotations of the image. Each element represents the presence (1) or absence (0) of a particular class. Due to the data imbalance, different weights are assigned to each class in action prediction: [1, 1, 2, 2]. These weights improve the flexibility of the model for learning small classes.

TABLE II
DETAILS OF ACTION AND REASON CATEGORIES IN THE BDD-OIA AND NU-AR DATASETS. THEY CONSIST OF FOUR ACTIONS AND TWENTY-ONE REASONS. THE DISTRIBUTION OF FRAMES IS PROVIDED SEPARATELY FOR EACH CATEGORY.

| | BDD-OIA | nu-AR | BDD-OIA | nu-AR | | BDD-OIA | nu-AR | BDD-OIA | nu-AR |
|---|---|---|---|---|---|---|---|---|---|
| Action Category | Number of samples for action categories | | F1 score for each action | | Reason Category | Number of samples for reason categories | | F1 score for each reason | |
| Move forward | 12,491 | 1,076 | 0.809 | 0.850 | Follow traffic | 7,805 | 351 | 0.664 | 0.519 |
| | | | | | Road is clear | 3,489 | 744 | 0.528 | 0.720 |
| | | | | | Traffic light is green | 4,838 | 268 | 0.537 | 0.403 |
| Stop/Slow down | 10,432 | 427 | 0.752 | 0.777 | Obstacle: car | 5,381 | 152 | 0.602 | 0.565 |
| | | | | | Obstacle: person/pedestrian | 1,539 | 108 | 0.509 | 0.610 |
| | | | | | Obstacle: rider | 233 | 10 | 0.114 | 0 |
| | | | | | Obstacle: others | 163 | 6 | 0 | 0 |
| | | | | | Traffic light | 5,255 | 346 | 0.795 | 0.627 |
| | | | | | Traffic sign | 455 | 2 | 0.250 | 0 |
| Turn left | 5,906 | 487 | 0.638 | 0.710 | Front car turning left | 150 | 4 | 0 | 0 |
| | | | | | On the left-turn lane | 666 | 3 | 0.215 | 0 |
| | | | | | Traffic light allows | 316 | 0 | 0.133 | 0 |
| Turn right | 6,532 | 497 | 0.627 | 0.536 | Front car turning right | 154 | 0 | 0 | 0 |
| | | | | | On the right-turn lane | 885 | 12 | 0.289 | 0.070 |
| | | | | | Traffic light allows | 365 | 5 | 0.121 | 0 |
| Can't change to left lane | - | - | - | - | Obstacles on the left lane | 4,503 | 221 | 0.588 | 0.525 |
| | | | | | No lane on the left | 4,514 | 471 | 0.509 | 0.655 |
| | | | | | Solid line on the left | 3,660 | 485 | 0.535 | 0.713 |
| Can't change to right lane | - | - | - | - | Obstacles on the right lane | 6,081 | 312 | 0.649 | 0.507 |
| | | | | | No lane on the right | 4,022 | 358 | 0.516 | 0.535 |
| | | | | | Solid line on the right | 2,161 | 478 | 0.513 | 0.832 |

For an end-to-end training of the proposed architecture, the model employs the Stochastic Gradient Descent (SGD) optimizer to update the weights and the learning rate scheduler dynamically adjusts the learning rate over time. The initial learning rate was set to 0.001, with a momentum of 0.9 and a weight decay of 1e-4. We utilized an NVIDIA A100 GPU on Google Colab. The model was trained for 50 epochs with the reduction ratio of 16.

### B. Datasets

This paper employs three different datasets for training, evaluating, and testing the proposed network. First, the BDD-10K dataset which is part of the BDD-100K dataset and contains 10,000 frames divided into 7,000 frames for training, 2,000 for testing, and 1,000 for validation which is used for semantic segmentation. Second, the BDD-OIA dataset which also utilizes a subset of the BDD-100K dataset, consists of 22,835 five-second videos. The final scene of each video is extracted into a separate file. From these, 22,835 annotated images are obtained and divided into 16,028 frames for training, 2,259 frames for validation, and 4,548 frames for testing. Third, the nuScenes Actions and Reasons (nu-AR) [45] dataset which contains 1,502 images from the nuTonomy scenes (nuScenes) [46] dataset. This dataset, similar to the BDD-OIA dataset, has action and reasoning annotations with 4 and 21 classes, respectively. The details of the names of each action and reason category, along with the number of images in each category for the BDD-OIA and nu-AR datasets, are presented in Tab. II.

### C. The Results

According to Tab. II, the number of images across different actions and reasons was not equally distributed, and the datasets were not balanced. Therefore, we used two metrics for evaluation: the overall F1 score and the mean F1 score.

Regarding quantitative evaluation, the results of the proposed model on the BDD-OIA dataset are presented in Tab. III, where the overall F1 score and mean F1 score metrics for actions and associated reasons are compared with the results of classical and state-of-the-art methods to show the superiority of the proposed model over other approaches. As shown in this table, our model outperforms all other methods in terms of explanation. Our model achieved the best performance for the explanation task among all models, with an overall F1 score of 0.551 and a mean F1 score of 0.384. The combination of attention-based feature refinement and multi-scale context aggregation allows the model to achieve better interpretability through attention maps. Furthermore, the hierarchical design for action and reason prediction and the corresponding feature fusion mechanism ensure consistency and logical coherence between action and reason predictions. The ResNet [34] model in Tab. III is the pre-trained ResNet101 model trained on the ImageNet [47] dataset and has a classifier, which serves as the baseline model. Unlike the Local Selector [48] approach, which focuses only on sparse object selection, our model provides a deeper understanding of the scene context by combining multi-scale features from different layers and simultaneously optimizing both channel and spatial attention. Compared to the CBM [49] approach, which requires manual concept labeling of the intermediate layers, our model is trained end-to-end without the need for intermediate concept labels, which reduces the cost and complexity of the data preparation. OIA uses a simple selector to select action-inducing objects, whereas our model combines multi-scale feature combination along with channel and spatial attention to significantly improve the diagnosis of action-inducing features. Unlike CBM-AUC [50], which learns concepts independently, our model better captures the causal dependencies between actions and reasons by connecting the reason head to the action outputs. Compared to NLE-DM [45], which only uses semantic segmentation, our attention-based multi-scale feature selection offers superior feature extraction capabilities at various scales through 7 different feature channels and information fusion from different layers. Compared to ABIM [51], which only focuses on the relationships between objects, our architecture combines both global and local information with a dual-attention mechanism and uses an auxiliary classifier to enhance learning. Finally, unlike the driver attention-based [52] model that uses a separate model for generating attention fusion, our model operates end-to-end, which prevents additional complexity.

TABLE III
COMPARISON OF THE RESULTS BETWEEN THE PROPOSED MODEL WITH A REDUCTION RATIO OF 16 AND CLASSIC AS WELL AS STATE-OF-THE-ART MODELS FOR ACTION AND REASON PREDICTION PERFORMANCE ON THE BDD-OIA DATASET. THE BEST RESULTS ARE SHOWN IN **BOLD**, AND THE NEXT-BEST RESULTS ARE IN *ITALICS*.

| Train/Validation/Test Set | BDD-OIA | | | |
|---|---|---|---|---|
| Networks | $F1_{mean}^{action}$ | $F1_{overall}^{action}$ | $F1_{mean}^{explanation}$ | $F1_{overall}^{explanation}$ |
| ResNet [34] | 0.392 | 0.601 | 0.180 | 0.331 |
| Local Selector [48] | 0.699 | 0.711 | 0.196 | 0.406 |
| CBM [49] | 0.610 | 0.661 | 0.292 | 0.412 |
| OIA [20] | *0.718* | **0.734** | 0.208 | 0.422 |
| CBM-AUC [50] | 0.658 | 0.704 | 0.342 | 0.522 |
| NLE-DM [45] | **0.723** | *0.733* | 0.312 | 0.517 |
| ABIM [51] | *0.718* | **0.734** | *0.335* | 0.537 |
| Driver attention-based explainable decision-making [52] | 0.593 | 0.695 | 0.267 | *0.538* |
| **Our mode ($\lambda$ =1)** | 0.707 | 0.710 | **0.384** | **0.551** |

The details of the results for each category of BDD-OIA and nu-AR datasets are listed in Tab. II. As can be seen in these tables, we also performed well on challenging classes, such as “turn left” and “turn right”. "Can't change to left/right lane" is not a part of the action categories; we included it symbolically to show the overall purpose of those 6 reasons, which appear across all of the other 4 categories.

TABLE IV
COMPARISON OF THE TEST RESULTS ON THE NU-AR DATASET BETWEEN THE PROPOSED MODEL WITH A REDUCTION RATIO OF 16 AND CLASSIC AS WELL AS STATE-OF-THE-ART MODELS FOR ACTION AND REASON PREDICTION PERFORMANCE. THE BEST RESULTS ARE SHOWN IN **BOLD**, AND THE NEXT-BEST RESULTS ARE IN *ITALICS*.

| Test Set | nu-AR | | | |
|---|---|---|---|---|
| Networks | $F1_{mean}^{action}$ | $F1_{overall}^{action}$ | $F1_{mean}^{explanation}$ | $F1_{overall}^{explanation}$ |
| OIA [20] | **0.738** | **0.777** | 0.216 | 0.396 |
| NLE-DM [45] | 0.688 | 0.722 | *0.308* | *0.499* |
| **Our model ($\lambda$ =1)** | *0.718* | *0.739* | **0.347** | **0.584** |

To demonstrate the generalization and robustness of our model, we also tested it on the nu-AR dataset described in the Datasets section. The results are compared with those of two classical and state-of-the-art models, as shown in Tab. IV. Our model outperformed the NLE-DM model in both the overall F1 score and mean F1 score metrics for reasoning by 8.5 and 3.9 percentage points, respectively. Additionally, our model outperformed the OIA model by 18.8 and 13.1 percentage points, respectively, for the same metrics on the test set.

The results of the proposed metric, called the joint F1 score, are also shown in Tab. V, which shows the superiority of our model in both metrics. Our model achieved 48.73% and 36.87% in the overall joint F1 score and mean joint F1 score, respectively, which fundamentally demonstrates the actual explainability and reliability of our model. This indicates that the model was able to make the correct decision for the correct reason in 48.73% of the cases. It should be noted that no code or implementation details are available for the other state-of-the-art models to reproduce their results on the nu-AR dataset or to evaluate them with the Joint F1 score.

TABLE V
COMPARISON OF RESULTS ON THE NEWLY PROPOSED METRICS OF OVERALL JOINT F1 SCORE AND MEAN JOINT F1 SCORE BETWEEN OUR MODEL AND THE STATE-OF-THE-ART NLE-DM METHOD ON THE BDD-OIA DATASET FOR A REDUCTION RATIO OF 16. THE BEST RESULTS ARE SHOWN IN **BOLD**, AND THE NEXT-BEST RESULTS ARE IN *ITALICS*.

| Test Set | BDD-OIA | |
|---|---|---|
| Networks | $F1_{mean}^{joint}$ | $F1_{overall}^{joint}$ |
| NLE-DM [45] | *0.319* | *0.462* |
| **Our model ($\lambda$ =1)** | **0.368** | **0.487** |

To demonstrate the superiority of the proposed method and provide a complete explanation of the black-box model, a visual explanation at the pixel level using the Grad-CAM method is shown in Fig. 3, which demonstrates the logical relationship between the attention of the model and important regions. Additionally, this illustrates a clear and direct relationship between the generated text and its visual explanation. Overall, Fig. 3 contains four rows; the first three rows show the results from the BDD-OIA dataset, and the 4th row shows the results from the nu-AR dataset. These rows demonstrate the following scenarios, respectively:

1. Correct action with correct reasoning:

The first row of Fig. 3 presents two images in which the model successfully and completely recognizes both the action and reasons at two different times of the day in two separate scenes with distinct actions and reasons. The original and visual explanation images are placed next to each other. These images highlight the key points of interest identified by the model and provide a visual representation that clearly displays the rationale behind the decision, which is perfectly aligned with the reasons selected for the prediction. This represents a reliable network of the study.

2. Correct action with wrong reasoning:

The right image in the second row of Fig. 3 shows an image in which the action is correctly detected, but the reasoning is incorrect. In this case, the model provides the reasoning "no lane on the right" for "turning right" because it highlighted the prohibition sign and the street guide, while it failed to understand the distance or mistakenly associated it with turning right. Meanwhile, the model did not perceive any reasoning for turning left and did not realize that it was in the left-turn lane. Therefore, an arbitrary decision was made.

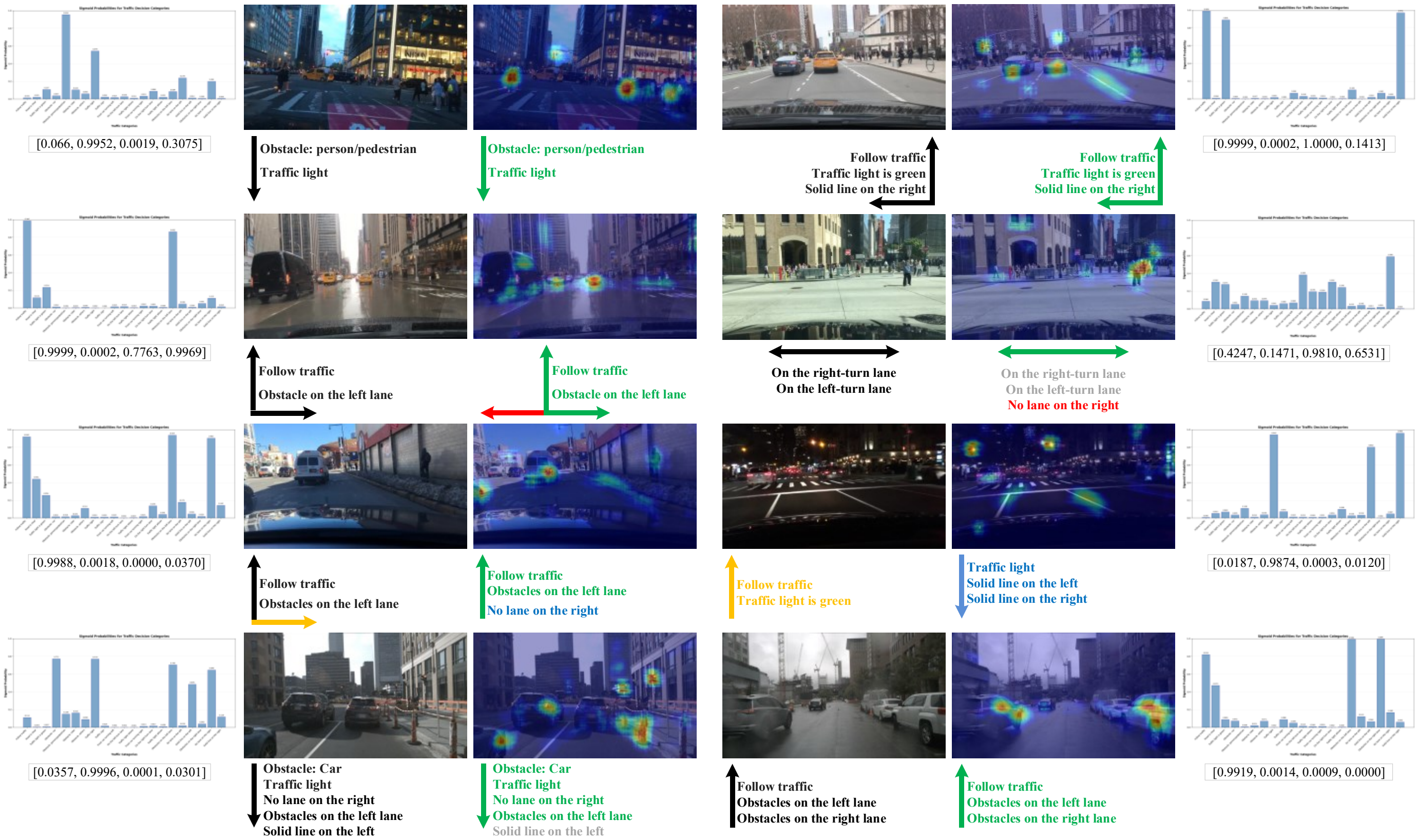

**Fig. 3.** Examples of visual and textual explanations along with sigmoid probabilities of actions and reasons under different weather and time conditions for the BDD-OIA and nu-AR test datasets. The first three rows of images are from the BDD-OIA dataset, while the last row of images is from the nu-AR dataset. The first row shows correct action-correct reason predictions, the second row shows examples of correct action-wrong reason and wrong action-correct reason predictions, and the third row shows cases where the human annotations are wrong, whereas the model correctly predicts the correct action-correct reason. The fourth row shows examples from the nu-AR dataset. Images from the dataset with human annotations are placed next to the visual explanations, along with their textual explanations as well as sigmoid probabilities of actions and reasons. Green, red, and gray colors indicate true positive, false positive and false negative results, respectively. The blue color indicates the correct diagnosis by the model, despite the wrong human annotations. The yellow color indicates the errors in the human annotations.

3. Wrong action with correct reasoning:

The second row of images on the left shows a case in which a wrong action is predicted, but the correct reason is identified. In this image, which depicts a rainy scene, although our proposed model correctly identifies that there is an obstacle on the left, the "turn left" action is predicted incorrectly. Additionally, the obstacle on the left is highlighted by Grad-CAM, which reveals that the model has focused on it, but a wrong decision has been made.

4. Wrong Human Annotations:

The human annotations were wrong, but the network itself behaved correctly, indicating the accuracy, strength, and robustness of the network. While the hand-labeled ground-truth incorrectly annotates the next move of the self-driving car as "turn right", there is "no lane on the right" side due to the presence of snow, a sidewalk, and a person. Our proposed model did not choose the wrong action and instead predicted the correct one. Furthermore, it also predicted the correct reason for not choosing to "turn right", which clearly demonstrates that the model made an accurate and reliable decision. In the picture on the right, both the action and reason annotated by the human are wrong, whereas our proposed model correctly predicts both the action and reason. However, these cases are not included in the F1 score calculation; otherwise, the metric values would have been significantly higher.

5. Images from the nu-AR dataset:

The bottom row of Fig. 3, displays two images from the nu-AR dataset, demonstrating that the model can also perform correctly on other datasets. In the image on the right, the model correctly and completely predicts the actions and reasons. In the image on the left, the model correctly predicts the actions and reasons, but fails to predict one of the reasons. This missed prediction had a sigmoid probability of 48.5%, which was too close to the threshold value to be selected.

Overall, the visual explanations demonstrate that the model has focused on several areas that were not included in the textual explanations due to classification limitations, and the model has achieved an excellent perception of the driving scene. The Grad-CAM images show that our model can effectively focus on and understand the key elements that are crucial for comprehending each driving scene.

### *D. Ablation Study*

To better understand the model performance, we conducted a comprehensive ablation study. Tab. VI shows the performance of the action and reasoning predictions for different $\lambda$ values. As explained in the Training Details section, $\lambda$, which is incorporated into the network loss function formula (Equation 22), is a parameter that can be used to control the relative importance of action and reasoning predictions. The best action result is obtained when $\lambda$ is set to 2, which means that the importance of learning the reason is twice that of the action. In the case where the importance of action is twice that of reason, that is, when $\lambda$=0.5, the second- best action results are obtained. Thus, the impact of learning from reason features is more than double the importance of action relative to the reason. When $\lambda$ is 0, we only have action, and the action outputs are stored for each sample. Although we might expect that, because it uses a single loss and only learns action, it would have the best action performance; in fact, it achieves the weakest performance in terms of the mean F1 score for action and is inferior to the other results. Therefore, multi-task learning has a small but positive impact on the action performance. However, the results of the four different $\lambda$ values that include action are not very different from each other, and the action output is relatively robust across different $\lambda$ values, regardless of whether reason prediction is present. This is because $\lambda$ has an impact of about 1% on the mean F1 score and about 0.5% on the overall F1 score for action prediction. Indeed, action predictions are not very sensitive to these variations, which is an advantage. To better evaluate the performance of the network, we conducted an experiment in which all the action inputs fed to the reasoning head were correct (represented as 0 and 1 values). We then trained the model with $\lambda$=∞, which corresponds to a network with only the reason head. In this case, the best results of reason were obtained, which achieved 60.77 for the overall F1 score and 40.14 for the mean F1 score. This demonstrates the significant impact that knowing the correct action from the beginning can have on the explanation performance. Moreover, we trained the model with $\lambda$ values of 0.5, 1, 2, and ∞ with our standard model architecture. In this architecture, we fed the sigmoid probabilities of the action for each image as inputs along with extracted features from previous layers to the reasoning module. When we fed the sigmoid probabilities during each epoch, initially, the action probabilities were not sufficiently correct; however, when feeding correct actions, the model received the same correct actions across all epochs. Consequently, knowing the correct decision can strongly contribute to generating an accurate explanation. The second-best reasoning results were achieved with $\lambda$=2. Additionally, in our previous study [53], we used $\lambda$=2, and the results improved by over 2% under the same configurations in this paper. A $\lambda$ of 0.5 produced the weakest reasoning results. When $\lambda$=∞, we only have the reasoning head, where the outputs obtained from the $\lambda$=0 were fed into the reasoning head, along with previously extracted features. In this case, an asynchronous training method was used primarily. Even the reasoning module alone, when given sigmoid probabilities, could not perform as well as the model with $\lambda$=2 with sigmoid probabilities. This occurred because we only fed the result from one epoch from the action-only network —which was from the best epoch of the network — into the reason-only network. Though, when both components were trained together, this happened dynamically across all epochs, which led to better network training. The presence of the action component also helped the reasoning module because when the reasoning module operated alone, its results were lower than those in all other cases, except when $\lambda$=0.5.

TABLE VI
AN ABLATION STUDY OF ACTION AND REASON PREDICTION PERFORMANCE, MEASURED BY THE OVERALL F1 SCORE AND MEAN F1 SCORE METRICS ON THE BDD-OIA DATASET, FOR DIFFERENT $\lambda$ (RELATIVE IMPORTANCE) VALUES (EQUATION 22) AND FOR FEEDING DIFFERENT ACTION OUTPUTS AS INPUTS TO THE REASONING HEAD. THE REDUCTION RATIO IS SET TO 16 FOR ALL EXPERIMENTS. THE BEST RESULTS ARE SHOWN IN **BOLD**, AND THE NEXT-BEST RESULTS ARE IN *ITALICS*.

| $\lambda$ | $F1_{mean}^{action}$ | $F1_{overall}^{action}$ | $F1_{mean}^{explanation}$ | $F1_{overall}^{explanation}$ |
|---|---|---|---|---|
| $\lambda = 0$ (only-action network) | 0.698 | 0.713 | - | - |
| $\lambda = 0.5$ (action-reason network with all outputs of action head in sigmoid probability format) | *0.706* | *0.714* | 0.351 | 0.538 |
| $\lambda = 1$ (action-reason network with all outputs of action head in sigmoid probability format) | *0.707* | 0.710 | 0.384 | 0.551 |
| $\lambda = 2$ (action-reason network with all outputs of action head in sigmoid probability format) | **0.709** | **0.715** | *0.387* | *0.563* |
| $\lambda = 2$ (Conference model [53]) | - | - | 0.366 | 0.540 |
| $\lambda = \infty$ (assumption: all actions are true) (only-reason network) | - | - | **0.401** | **0.607** |
| $\lambda = \infty$ (only-reason network with actions from only-action network in sigmoid probability format) | - | - | 0.353 | 0.542 |

Tab. VII shows a comparison of the prediction performance for the action and reason across different reduction ratios. As explained in The Proposed Model section, the reduction ratio is a parameter that balances the trade-off between accuracy and complexity by reducing the number of parameters in the model. In the most complex case, the reduction ratio adds approximately 526,000 parameters to our model, which accounts for about 1% of the total model complexity. Therefore, we do not consider the computational complexity associated with the reduction ratio to be a limiting factor in our study and focus on examining the accuracy of the model at different reduction ratios. The best action performance was achieved at a reduction ratio of 4. The second-best results for action occurred at reduction ratios of 1 and 32. The best performance for reason occurred at a reduction ratio of 16, and the second-best result for reason occurred at a reduction ratio of 4. Notably, the results are very close. The reduction ratio for action had an effect of approximately 0.5 percentage points on the mean F1 score and more than 2 percentage points on the overall F1 score. Meanwhile, for reason, it had an effect of 2 percentage points on the mean F1 score and fewer than 0.5 percentage points on the overall F1 score. As shown in Tab. VII, there is no linear relationship between the reduction ratio and F1 score performance. Overall, it appears that having an equal number of features in the channel attention neural network may not be optimal, and a reduction ratio of 4 or 16,

TABLE VII
AN ABLATION STUDY OF ACTION AND REASON PREDICTION PERFORMANCE, MEASURED BY THE OVERALL F1 SCORE AND MEAN F1 SCORE METRICS ON THE BDD-OIA DATASET, FOR DIFFERENT REDUCTION RATIOS IN THE ATTENTION MODULE. IN ALL EXPERIMENTS, $\lambda$ IS EQUAL TO 1. THE BEST RESULTS ARE SHOWN IN **BOLD**, AND THE NEXT-BEST RESULTS ARE IN *ITALICS*.

| Reduction ratio | $F1_{mean}^{action}$ | $F1_{overall}^{action}$ | $F1_{mean}^{explanation}$ | $F1_{overall}^{explanation}$ | Number of Parameters | Number of CBAM-DCA-ResNet50 parameters |
|---|---|---|---|---|---|---|
| 1 | *0.707* | 0.717 | 0.364 | 0.547 | 49.24M | 526,444 |
| 4 | **0.710** | **0.721** | *0.379* | *0.550* | 48.84M | 132,844 |
| 16 | 0.707 | 0.710 | **0.384** | **0.551** | 48.75M | 34,444 |
| 32 | 0.705 | *0.719* | 0.371 | 0.548 | 48.73M | 18,044 |

which is moderate for this model, proves to be beneficial.

## VI. CONCLUSION

In this study, we addressed the problems of explainable decision-making in autonomous driving. We presented a novel end-to-end model that overcomes the limitations of previous models, which is grounded in an attention-based multi-scale feature extractor that employs multitasking and feeds the action output to the reasoning head to produce appropriate explanations. We introduced a new metric, the Joint F1 score, which performs real explainability assessments of autonomous vehicles. The generalization of our work on the nu-AR dataset was also demonstrated by achieving higher explanation accuracy than other models. The visual explanation also helped to make deep learning models in autonomous driving more explainable, which were both logical and consistent with the text output, revealing that the model has a complete and comprehensive understanding of the driving scenes. Additionally, a comprehensive ablation study was performed. This work makes a significant contribution to advancing the understanding of explainability and developing safer and more reliable deep learning systems for autonomous driving.

In future work, we plan to reprocess the dataset to address its issues, including relabeling, class balancing, and selecting more diverse and appropriate labels. Furthermore, we aim to enhance the reliability of the system to ensure that it provides correct explanations for a greater proportion of decisions. Finally, we suggest that in future datasets, annotations could be labeled as joint (action, reason) pairs instead of two separate lists, as such pairing could constitute more realistic annotations for rigorous evaluation of action-explanation alignment in autonomous driving.

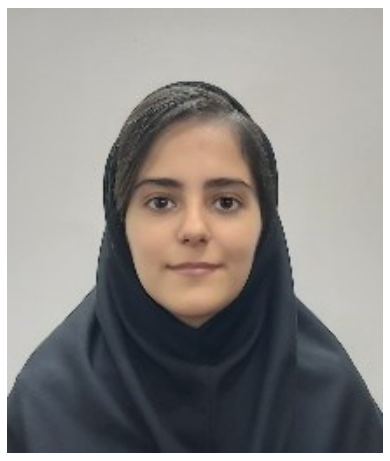

**Maryam Sadat Hosseini Azad** received her B.Sc. degree in Electrical and Electronic Engineering from Qom University of Technology, Qom, Iran, in 2022, and M.Sc. in Electrical and Digital Electronic Systems Engineering from Iran University of Science & Technology, Tehran, Iran, in 2025. She has been a research assistant and member of the Computer Vision Lab at the Department of Electrical Engineering at Iran University of Science & Technology since 2022. Her research interests include computer vision, explainable artificial intelligence (XAI), and autonomous driving.

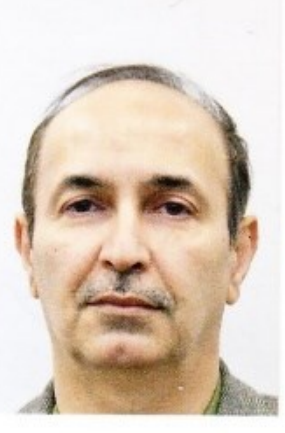

**Shahriar B. Shokouhi** received his B.Sc. and M.Sc. degrees in Electrical and Electronic Engineering from Iran University of Science & Technology in 1986 and 1989, respectively. He received his Ph.D. degree in Electronic and Electrical Engineering from the University of Bath, UK in 1999. In July 1999, he joined the School of Electrical Engineering at Iran University of Science and Technology in Tehran, Iran, where he is currently emeritus faculty member. He is a member of the electronics group and director of the optoelectronics and machine vision research laboratories. His research interests include machine vision systems design, trusted hardware design, and intelligent systems design.

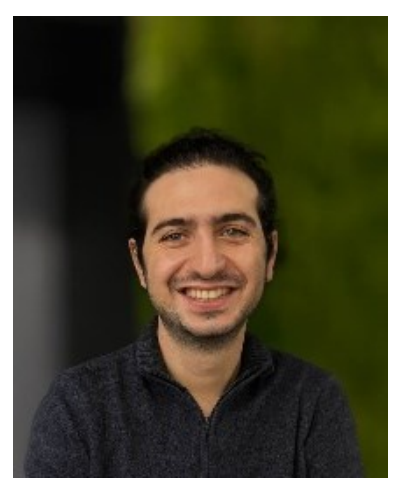

**Amir Abbas Hamidi Imani** received his bachelor's and master's degrees in Electrical Engineering from Shahid Beheshti University, where he worked on FPGA-based Sobel edge detection and vehicle tracking in videos. He is currently pursuing his Ph.D. at Iran University of Science and Technology, focusing on robust object tracking to improve the reliability of visual recognition systems in real-world environments. His research interests include computer vision, natural language processing, and applications of artificial intelligence in healthcare.

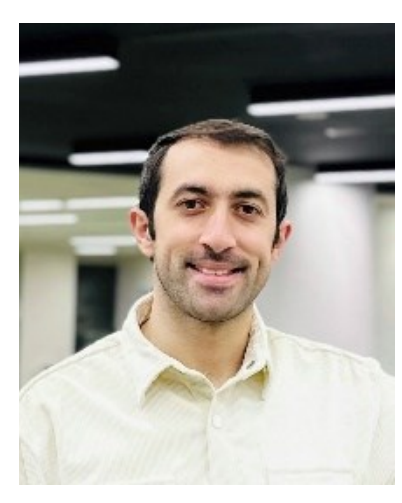

**Shahin Atakishiyev** (Member, IEEE) received the B.Sc. degree in computer engineering from Qafqaz University, Azerbaijan, in June 2015, the M.Sc. degree in software engineering and intelligent systems from the University of Alberta, Canada, in January 2018, and the Ph.D. degree in computing science in September 2024. During his Ph.D., he built explainable artificial intelligence (XAI) approaches for autonomous vehicles under the supervision of Prof. Randy Goebel, and his thesis received the 2024 Outstanding PhD Thesis Runner-Up Award at the University of Alberta. Shahin's research interests include developing safe, explainable, ethical, and human-centered artificial intelligence approaches for real-world problems.

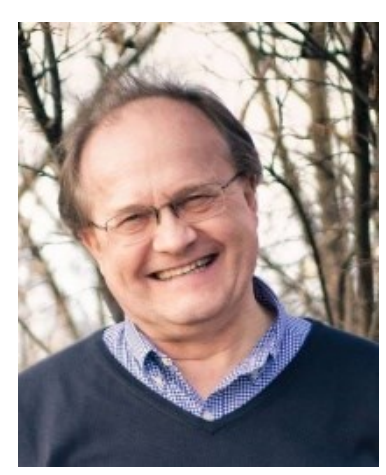

**Randy Goebel** is a Professor of Computing Science and adjunct Professor in the Faculty of Medicine at the University of Alberta, and Fellow and Co-founder of the Alberta Machine Intelligence Institute (Amii), one of three Canadian federally-funded AI research organizations. He has had faculty appointments and visiting faculty appointments at the University of Waterloo, University of Regina, University of Tokyo, Hokkaido University (Sapporo, Japan), Multimedia University (Kuala Lumpur, Malaysia), Instituto Tecnológico de Monterrey (Monterrey, Mexico), and has been a visiting researcher at the German Center for AI Research (DFKI), the National Institute for Informatics (NII, Tokyo), and the Volkswagen Data Lab (Munich). His research interests include formal knowledge representation and reasoning (induction, belief revision, explainable AI (XAI)), knowledge visualization, algorithmic complexity, natural language processing (NLP), systems biology, with applications in clinical medicine, legal reasoning, and automated driving.